\documentclass[conference]{IEEEtran}
\ifCLASSINFOpdf
  \usepackage[pdftex]{graphicx}
\else
\fi
\usepackage{amsmath,amsfonts}
\usepackage{cite}
\usepackage{listings}
\usepackage{flushend}

\usepackage{amsmath}
\usepackage{amsfonts} 
\usepackage{amsthm}
\usepackage{amssymb}
\usepackage[boxruled]{algorithm2e}
\usepackage{amsthm} 
\usepackage{mathtools} 

\usepackage{BOONDOX-ds}

\usepackage{bbm}

\begin{document}
%
\title{Sparse Deep Neural Network Exact Solutions}

\author{\IEEEauthorblockN{
  Jeremy Kepner$^1$,
  Vijay Gadepally$^1$,
  Hayden Jananthan$^{1,2}$,
  Lauren Milechin$^1$,
  Sid Samsi$^1$ 
}
\vspace{1ex}
\IEEEauthorblockA{
	$^1$Massachusetts Institute of Technology, $^3$Vanderbilt University}
}


%


\pagestyle{plain}

\maketitle

\begin{abstract}
Deep neural networks (DNNs) have emerged as key enablers of machine learning. Applying larger DNNs to more diverse applications is an important challenge. The computations performed during DNN training and inference are dominated by operations on the weight matrices describing the DNN.  As DNNs incorporate more layers and more
neurons per layers, these weight matrices may be required to be sparse because of memory limitations.  Sparse DNNs are one possible approach, but the underlying theory is in the early stages of development and presents a number of challenges, including determining the accuracy of inference and selecting nonzero weights for training.   Associative array algebra has been developed by the big data community to combine and extend database, matrix, and graph/network concepts for use in large, sparse data problems. Applying this mathematics to DNNs simplifies the formulation of DNN mathematics and reveals that DNNs are linear over oscillating semirings.  This work uses associative array DNNs to construct exact solutions and corresponding perturbation models to the rectified linear unit (ReLU) DNN equations that can be used to construct test vectors for sparse DNN implementations over various precisions.  These solutions can be used for DNN verification, theoretical explorations of DNN properties, and a starting point for the challenge of sparse training.
\end{abstract}


%
\IEEEpeerreviewmaketitle

\thispagestyle{plain}

\section{Introduction}
\let\thefootnote\relax\footnotetext{This material is based in part upon
work supported by the NSF under grant number DMS-1312831.  Any opinions,
findings, and conclusions or recommendations expressed in this material
are those of the authors and do not necessarily reflect the views of
the National Science Foundation.}
Machine learning has been the foundation of artificial intelligence since its inception
\cite{ware1955introduction,clark1955generalization,selfridge1955pattern,dinneen1955programming,newell1955chess,mccarthy2006proposal,minsky1960learning,minsky1961steps}. Standard machine learning applications include speech recognition \cite{selfridge1955pattern}, computer vision \cite{dinneen1955programming}, and even board games \cite{newell1955chess,samuel1959some}.

\begin{figure}[htb]
  	\centering
    	\includegraphics[width=\columnwidth]{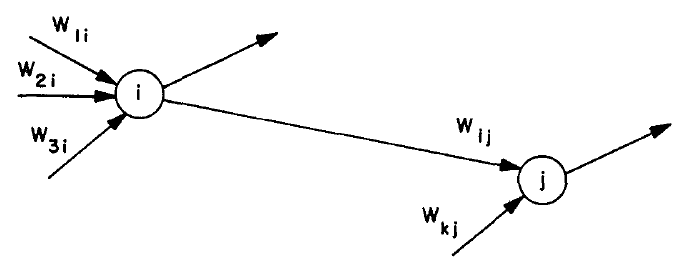}
      	\caption{Typical network elements $i$ and $j$ showing connection weights $w$ (reproduced from  \cite{clark1955generalization})}
      	\label{fig:clark1955fig1}
\end{figure}

\begin{figure}[htb]
  	\centering
    	\includegraphics[width=\columnwidth]{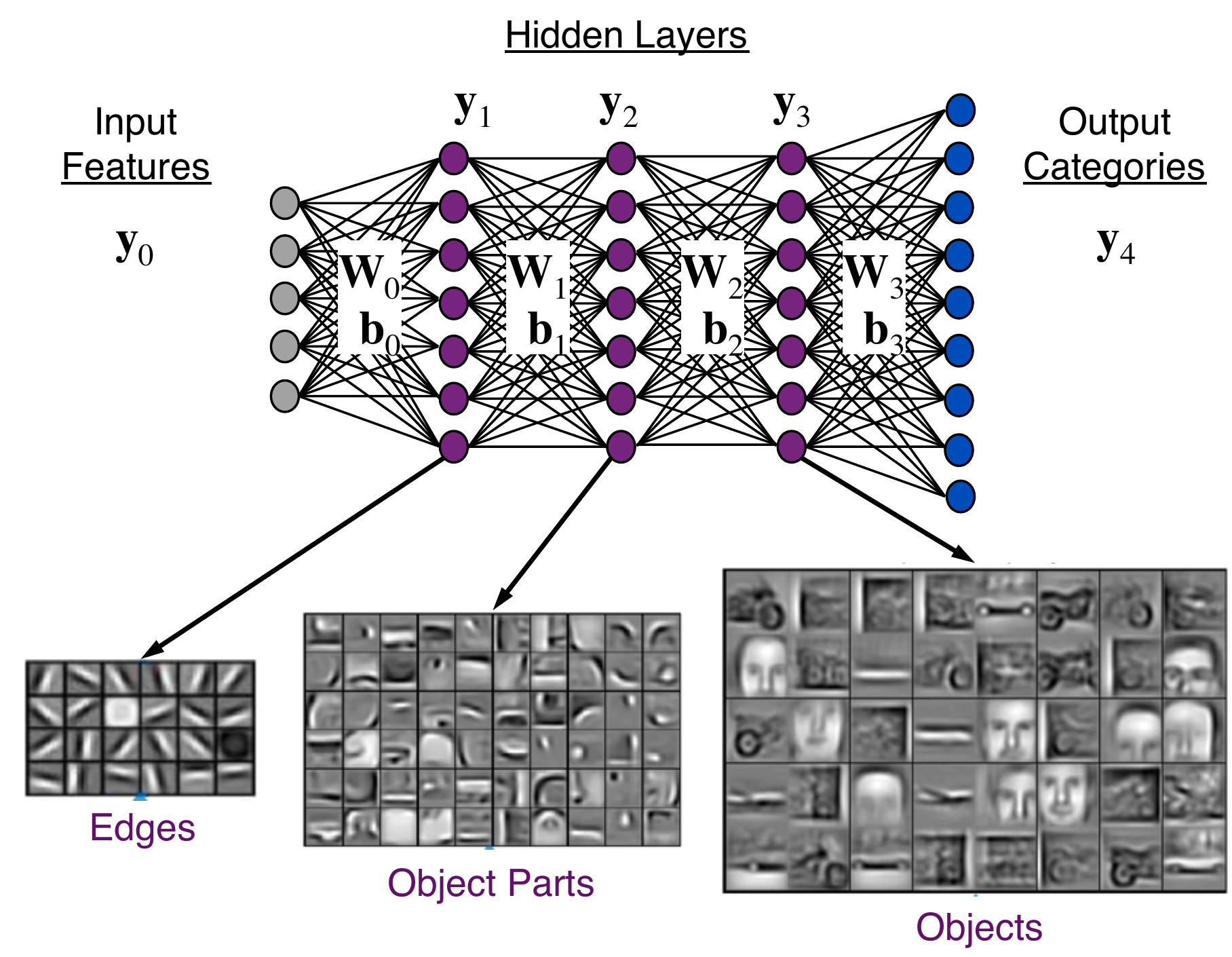}
	\caption{Four layer ($L=4$) deep neural network architecture
	for categorizing images.  The input
	features  ${\bf y}_0$ of an image are passed through a series
	of network layers ${\bf W}_{\ell=0,1,2,3}$, with bias terms
	${\bf b}_{\ell=0,1,2,3}$, that produce scores for categories
	${\bf y}_{L=4}$.  (Figure adapted from \cite{lee2009convolutional})}
      	\label{fig:DNNarchitecture}
\end{figure}

Drawing inspiration from biological neurons to implement machine learning was the topic of the first paper presented at the first machine learning conference in 1955 \cite{ware1955introduction,clark1955generalization} (see Figure~\ref{fig:clark1955fig1}). It was recognized very early on in the field that direct computational training of neural networks was computationally unfeasible with the computers that were available at that time \cite{minsky1960learning}.  The many-fold improvement in neural network computation and theory has made it possible to create neural networks capable of better-than-human performance in a variety of domains \cite{lippmann1987introduction,reynolds2000speaker,krizhevsky2012imagenet,lecun2015deep}. The production of validated data sets \cite{campbell1995testing,lecun1998mnist,deng2009imagenet} and the power of graphic processing units (GPUs) \cite{campbell2002deep,mcgraw2007benchmarking,kerr2008gpu,epstein2012making}
have allowed the effective training of deep neural networks (DNNs) with 100,000s of input features, $N$, and 100s of layers, $L$, that are capable of choosing from among 100,000s categories, $M$ (see Figure~\ref{fig:DNNarchitecture}).

The impressive performance of large DNNs provides motivation to explore even larger networks.  However, increasing $N$, $L$, $M$. each by a factor 10 results in a 1000-fold increase in the memory required for a DNN.  Because of these memory constraints, trade-offs are currently being made in terms of precision and accuracy to save storage and computation \cite{liu2015sparse,lavin2016fast,jouppi2017datacenter,kepner2017enabling}. Thus, there is significant interest in exploring the effectiveness of sparse DNN representations where many of the weight values are zero.  As a comparison, the human brain has approximately 86 billion neurons and 150 trillion synapses~\cite{CNE:CNE21974}.  Its graph representation would have approximately 2,000 edges per node, or a sparsity of $2 \times 10^3 / 86 \times 10^9 = 0.000002\%$.

If a large fraction of the DNN weights can be set to zero, storage and computation costs can be reduced proportionately \cite{iandola2016squeezenet,shi2017speeding}.  The interest in sparse DNNs is not limited to their computational advantages. There has also been extensive theoretical work exploring the potential neuromorphic and algorithmic benefits of sparsity \cite{lee2008sparse,boureau2008sparse,glorot2011deep,DBLP:journals/corr/HanMD15,yu2012exploiting}.

Theoretical explorations of dense DNNs are receiving renewed attention \cite{saxe2013exact,choromanska2015open,liao2017theory,zhang2017theory}. The underlying theory of sparse DNNs is in the early stages of development and presents a number of challenges, including verification (determining the accuracy errors of a sparse network) and training (selecting nonzero weights).   Associative array algebra has been developed by the big data community to combine and extend database, matrix, and graph/network concepts for use in large, sparse data problems. Applying this mathematics to DNNs simplifies the formulation of DNN mathematics, unifies broad classes of DNNs, and reveals that DNNs are linear over oscillating semirings.  This work provides a methodology for constructing exact solutions and corresponding perturbation models to the rectified linear unit (ReLU) DNN equations that can be used to construct test vectors for sparse DNN implementations over various precisions.  These solutions can be used for DNN verification, theoretical explorations of DNN properties, and a starting point for the  challenge of sparse training.

\begin{figure*}[ht]
  	\centering
    	\includegraphics[width=7in]{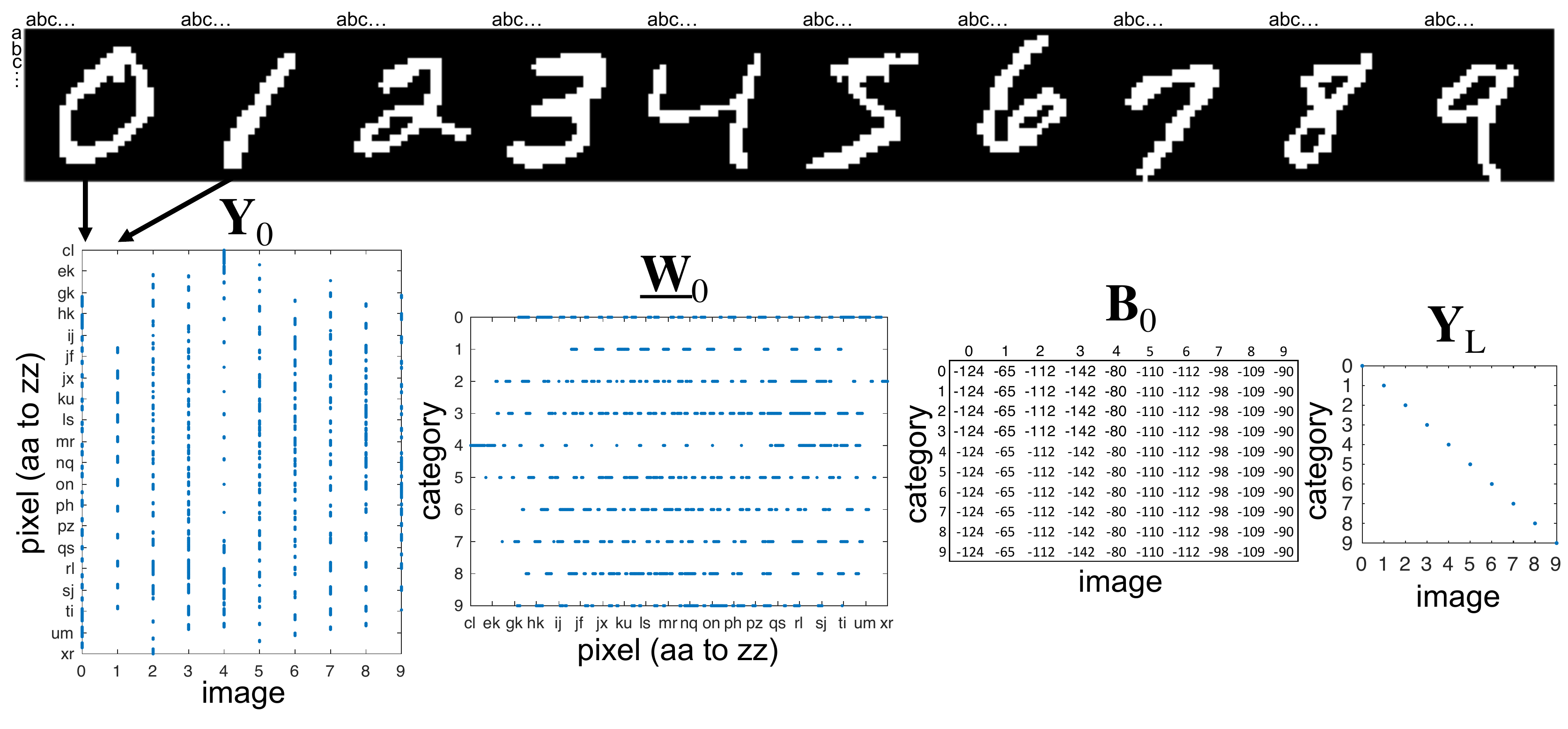}
	\caption{Example exact solution using MNIST data. The $28 \times 28$ MNIST images are trimmed to $26 \times 26$ so that each pixel can be indexed with two lowercase letters.  The images are thresholded at a value of 0.5 so that each pixel is either 0 or 1.  The nonzero pixels are then used to form the single layer exact solution $\underline{\mathbf{W}}_0$ and corresponding input $\mathbf{Y}_0$.  The bias values $\mathbf{b}_0$, corresponding to any row of $\mathbf{B}_0$, are then chosen so that each input only selects one category as output in $\mathbf{Y}_L$.}
      	\label{fig:MNIST}
\end{figure*}

\section{DNN Associative Array Formulation}

   Associative array algebra simplifies combining large numbers of solutions to DNN equations with varying layers and neurons.  The mathematics of associative arrays and how they encompass databases, matrices, graphs, and networks is fully described in the text \cite{kepner2018mathematics}.  Only the essential mathematical properties of associative arrays necessary for describing DNNs are reviewed here and in Appendix A.  In brief, an associative array $\mathbf{A}$ is defined as a mapping from sets of keys to values
$$
  \mathbf{A}: K_1 \times K_2 \to \mathbb{V}
$$
where $K_1$ are the row keys and $K_2$ are the column keys and can be any sortable set, such as integers, real numbers, and strings. In a DNN, the row keys and column can be used to clearly label features, neurons, batches, and categories.  $\mathbb{V}$ is the set of values that form a semiring $(\mathbb{V},\oplus,\otimes,0,1)$ with addition operation $\oplus$, multiplication operation $\otimes$, additive identity/multiplicative annihilator 0, and multiplicative identity 1.  Most significantly, the properties of associative arrays are determined by the properties of the value set $\mathbb{V}$.  In other words, if $\mathbb{V}$ is linear (distributive), then so are the corresponding associative arrays.  Associative arrays allow arbitrary key labels for neurons which greatly simplifies the bookkeeping of DNN solutions, and so the number of neurons in a layer and even the number of layers need not be specified.

  The primary mathematical operation performed by a DNN network is the inference, or forward propagation, step.  Inference is executed repeatedly during training to determine both the weight arrays ${\bf W}_\ell$ and the bias vectors ${\bf b}_\ell$ of the DNN.  The inference computation shown in Figure~\ref{fig:DNNarchitecture} is given by
$$
  {\bf y}_{\ell + 1} = h({\bf W}_\ell {\bf y}_\ell + {\bf b}_\ell)
$$
where $h()$ is a nonlinear function applied to each element of the vector.  A commonly used function is the rectified linear unit (ReLU) given by
$$
   h({\bf y}) = \max({\bf y},0)
$$
which sets values less that 0 to 0 and leaves other values unchanged.  When training a DNN, it is usually necessary to compute multiple ${\bf y}_\ell$ vectors at once in a batch that can be denoted as the array ${\bf Y}_\ell$.  In associative array form, the inference step becomes
$$
  {\bf Y}_{\ell + 1} = h({\bf W}_\ell {\bf Y}_\ell + {\bf B}_\ell)
$$
where ${\bf B}_\ell$ is a replication of ${\bf b}_\ell$ along columns given by
$$
  \mathbf{B}_\ell = \mathbf{b}_\ell |\mathbf{Y}_\ell \mathbf{1}|_0
$$
and $\mathbf{1}$ is a column array of 1's, and $| ~ |_0$ is the zero norm.

  If $h()$ were a linear function, then the above equation could be solved exactly and the computation could be greatly simplified.  However, current evidence suggests that the nonlinearity of $h()$ is required for a DNN to be effective.  Interestingly, the inference computation can be rewritten as a linear function over two different semirings
$$
  {\bf y}_{\ell + 1} ={\bf W}_\ell {\bf y}_\ell \otimes {\bf b}_\ell \oplus 0
$$
or in array form
$$
  {\bf Y}_{\ell + 1} ={\bf W}_\ell {\bf Y}_\ell \otimes {\bf B}_\ell \oplus 0
$$
where the $\oplus = \max$ and $\otimes = +$.  Thus, ${\bf W}_\ell {\bf y}_\ell$ and ${\bf W}_\ell {\bf Y}_\ell$ are computed over the standard arithmetic ${+}.{\times}$ semiring 
$$
  S_1 = (\mathbb{R},+,\times,0,1)
$$
while the $\oplus$ and $\otimes$ operations are performed over the ${\max}.{+}$ semiring
$$
  S_2 = (\{ \text{-}\infty \cup \mathbb{R} \},\max,+,\text{-}\infty,0)
$$
Thus, the ReLU DNN can be written as a linear system that oscillates over two semirings $S_1$ and $S_2$.  $S_1$ is the most widely used of semirings and performs standard correlation between vectors.  $S_2$ is also a commonly used semiring for selecting optimal paths in graphs.  Thus, the inference step of a ReLU DNN can be viewed as combining correlations of inputs to choose optimal paths  through the neural network. Furthermore, by uniquely labeling features/neurons in the associative arrays, the entire architecture can be collapsed to single arrays
$$
   \mathbf{W} = \bigoplus_\ell \mathbf{W}_\ell  ~~~~~~~~~~~~~~~~~~ \mathbf{B} = \bigoplus_\ell \mathbf{B}_\ell
$$
resulting in the following recursive DNN equation that is iterated $L$ times
$$
  {\bf Y} \leftarrow {\bf W} {\bf Y} \otimes {\bf B} \oplus 0
$$

\section{Model Problem}

  While general analytic solutions to the DNN equations are currently not known, it is possible to simplify DNN problems so that such solutions can be constructed.  First, constrain all weight arrays $\mathbf{W}_\ell$ to have values of either 0 or 1.  Next, let $\underline{\mathbf{W}}_0$ be a single layer $L=1$ exact solution to a DNN problem.  Another DNN will be an exact solution if setting the input to
$$
  \mathbf{Y}_0 = \underline{\mathbf{W}}^{\sf T}_0
$$
results in an output
$$
  \mathbf{Y}_L = \mathbb{I}
$$
where $\mathbb{I}$ is the identity array (see Appendix A).  An example of such a construction using MNIST data \cite{lecun1998mnist} is shown in Figure~\ref{fig:MNIST}.  To construct this example, the $28 \times 28$ MNIST images are trimmed to $26 \times 26$ so that each pixel can be indexed with two lowercase letters.  In addition, the images are thresholded at a value of 0.5 so that each pixel is either 0 or 1.  The nonzero pixels are then used to form the single layer exact solution $\underline{\mathbf{W}}_0$ and corresponding input $\mathbf{Y}_0$.  The bias values $\mathbf{b}_0$, corresponding to any row of $\mathbf{B}_0$, are then chosen so that each input only selects one category as output in $\mathbf{Y}_L$ using the formula
$$
\mathbf{b}_\ell = \beta_\ell | \mathbf{W}_\ell \mathbf{1} |_0 - \mathbf{W}_\ell \mathbf{1}
$$
where $0 < \beta_\ell \leq 1$ (assume $\beta_\ell = 1$ unless stated otherwise).  While the above formulated DNN exact solutions are unlikely to produce low error rates when applied to the full MNIST data set, they do capture the sparsity structure of the data that can be used for verifying real DNN systems and for theoretical studies of DNNs.

  The above MNIST problem can be used to generate a wide range of exact DNN solutions, but their scale is difficult to visualize.  Without loss of generality, two mechanisms for generating DNN exact solutions will be provided for arbitrary feature labels for the simpler case of using a DNN to select common words.

\section{Combinatoric Construction}

  Exact DNN solutions derived from real data are extremely useful.  It is also useful to generate arbitrary DNN solutions to explore a wider range of DNN architectures.  The combinatoric construction builds an exact DNN solution using  Kronecker products of associative arrays of features.  These solutions have a category for all possible feature combinations and can be viewed as a superset of any real DNN.  These solutions do not require real data to construct, and they provide for exponential growth in output as features are added.
  
\begin{figure}[htb]
  	\centering
    	\includegraphics[width=\columnwidth]{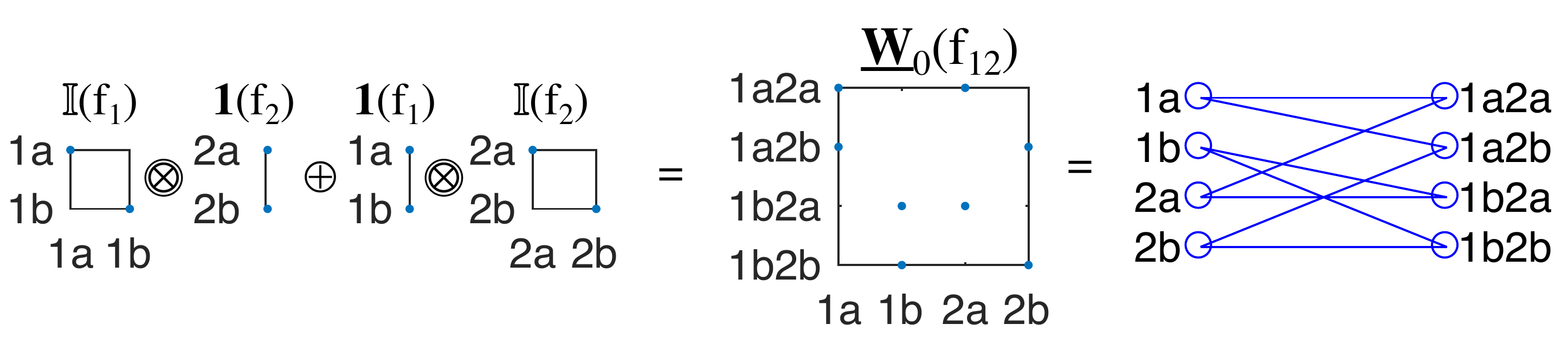}
	\caption{Construction of a single layer exact solution $\underline{\mathbf{W}}_0({\rm f}_{12})$ combining  features f$_1$ and f$_2$.}
      	\label{fig:Combinatoric1stLayer}
\end{figure}

\begin{figure}[htb]
  	\centering
    	\includegraphics[width=\columnwidth]{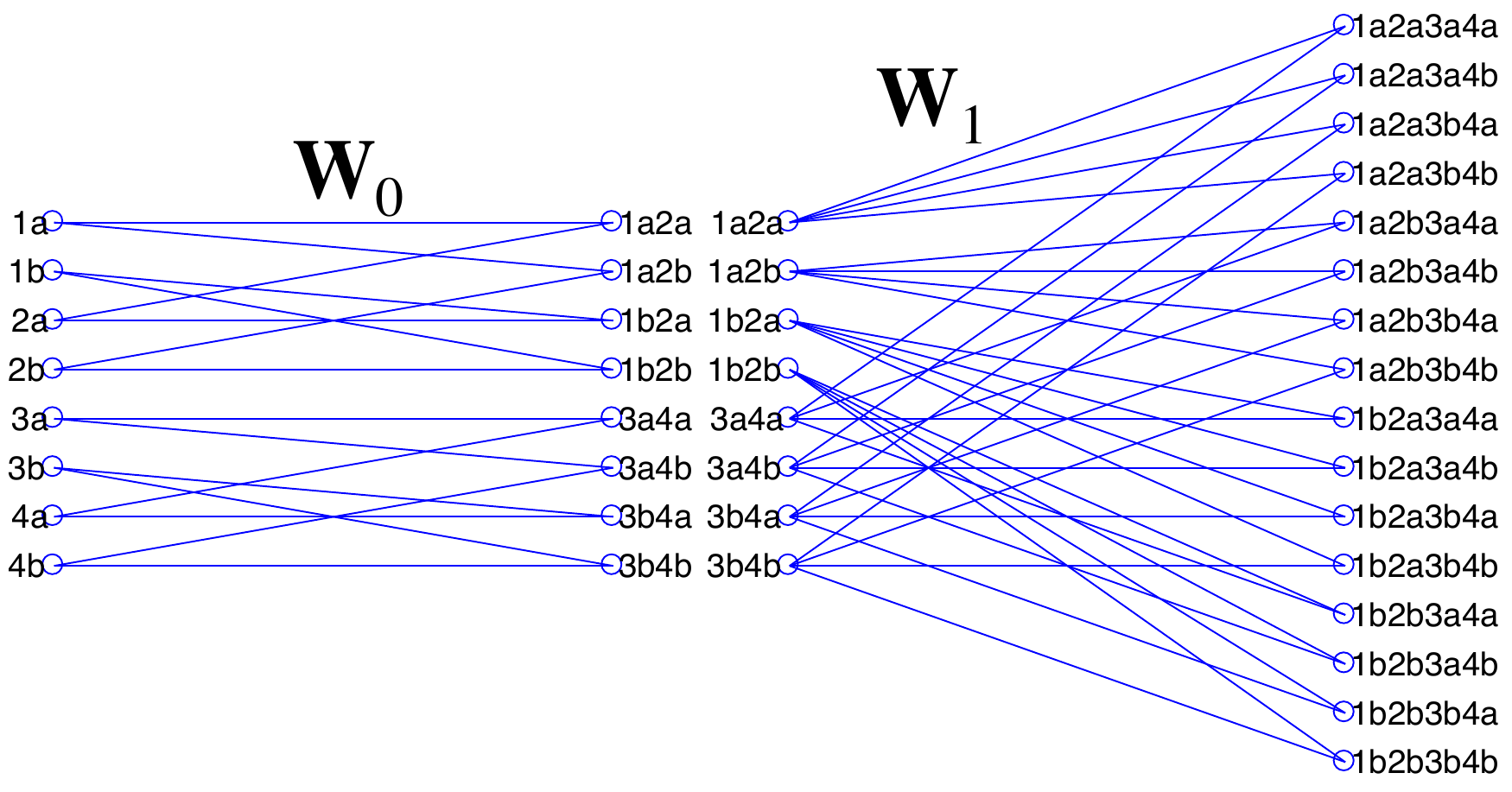}
\newline
\newline
	    \includegraphics[width=\columnwidth]{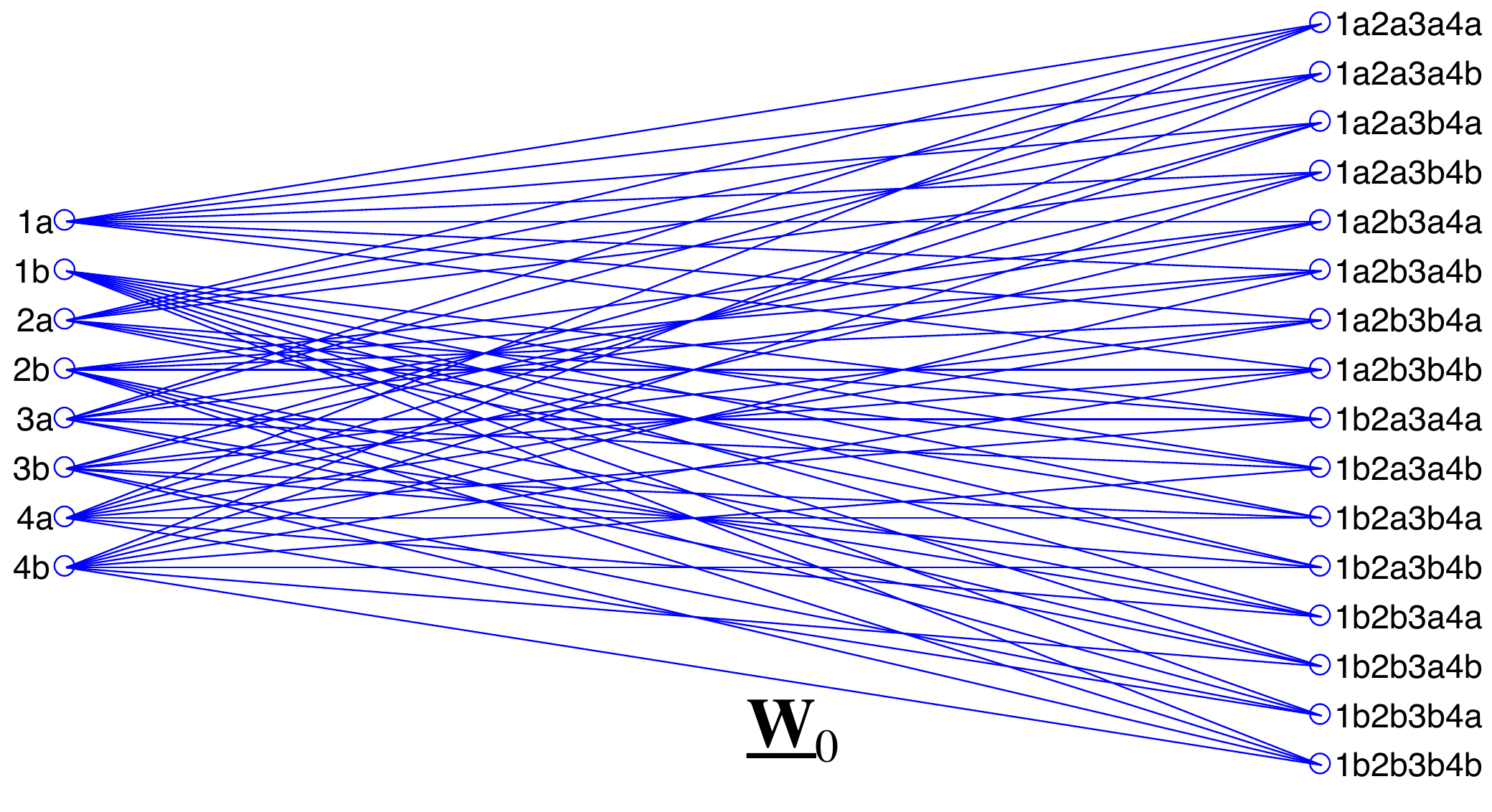}
	\caption{(top) Two layer exact DNN solution $\mathbf{W}_0$ and $\mathbf{W}_1$ over feature sets ${\rm f}_1 =\{{\rm 1a, 1b}\}$, ${\rm f}_2 =\{{\rm 2a, 2b}\}$, ${\rm f}_3 =\{{\rm 3a, 3b}\}$, and ${\rm f}_4 =\{{\rm 4a, 4b}\}$. (bottom) Single layer exact DNN solution $\underline{\mathbf{W}}_0$ over the same feature sets.}
      	\label{fig:Combinatoric1st2ndLayer}
\end{figure}

    Consider distinctly labeled feature sets
$$
 {\rm f}_1 =\{{\rm 1a, 1b}\} ~~~~ {\rm f}_2 =\{{\rm 2a, 2b}\} ~~~~
 {\rm f}_3 =\{{\rm 3a, 3b}\} ~~~~ {\rm f}_4 =\{{\rm 4a, 4b}\}
$$
A single layer exact solution $\underline{\mathbf{W}}_0({\rm f}_{12})$ that combines features f$_1$ and f$_2$ can be constructed via the Kronecker product \text{\textcircled{$\otimes$}} as shown in Figure~\ref{fig:Combinatoric1stLayer}.  The weight array $\underline{\mathbf{W}}_0({\rm f}_{34})$ can be constructed through similar means.  The first layer of a two layer exact solution can then be computed as
$$
  \mathbf{W}_0 = \underline{\mathbf{W}}_0({\rm f}_{12}) \oplus \underline{\mathbf{W}}_0({\rm f}_{34})
$$
The second layer can be computed in a manner similar to the first layer via
$$
  \mathbf{W}_1 = \mathbb{I}({\rm f}_{12}) \text{\textcircled{$\otimes$}} \mathbf{1}({\rm f}_{34}) ~ \oplus ~
                 \mathbf{1}({\rm f}_{34}) \text{\textcircled{$\otimes$}} \mathbb{I}({\rm f}_{12}) 
$$
The resulting two layer exact DNN solution is depicted in Figure~\ref{fig:Combinatoric1st2ndLayer} (top).
The complete single layer exact solution shown in Figure~\ref{fig:Combinatoric1st2ndLayer} (bottom) can be computed by
$$
  \underline{\mathbf{W}}_0 = \mathbf{W}_1\mathbf{W}_0 
$$
where values greater than 1 are set to 1 using the zero norm $| ~ |_0$ or by performing the above array multiplication using ${\max}.{\times}$ instead of ${+}.{\times}$.  The unique labeling of features/neurons in the associative arrays allows the entire architecture to be collapsed into the single arrays
$$
   \mathbf{W} = \mathbf{W}_0 \oplus \mathbf{W}_1 ~~~~~~~~~~~~ \mathbf{B} = \mathbf{B}_0 \oplus \mathbf{B}_1
$$
This combinatorial approach can be scaled arbitrarily over feature sets to produce a large number of DNN exact solutions over a wide range of DNN architectures.

\section{Selective Construction}

Given a set of features, such as f$_1$ and f$_2$, and known categories f$_{12}$, it is possible to construct an exact DNN solution by selecting only those features that contribute to known categories.  Consider as target categories the ordered list of popular two letter words

f$_{12}$ = (ad,ah,am,as,at,be,by,do,go,ha,he,hi,ie,if,in,it,me,

~~~~~~~ mr,ms,my,no,of,oh,ok,on,or,pc,pm,re,so,to,tv,uh,

~~~~~~~ up,us,vs,we)

\noindent These categories select the input feature ordered lists

f$_1$ = (a,b,d,g,h,i,m,n,o,p,r,s,t,u,v,w)

f$_2$ = (a,c,d,e,f,h,i,k,m,n,o,p,r,s,t,v,y)

\noindent Note: the feature identifiers f$_1$ = (1a,1b, \ldots ) are still present but are omitted for readability.

\begin{figure}[htb]
  	\centering
	    \includegraphics[width=\columnwidth]{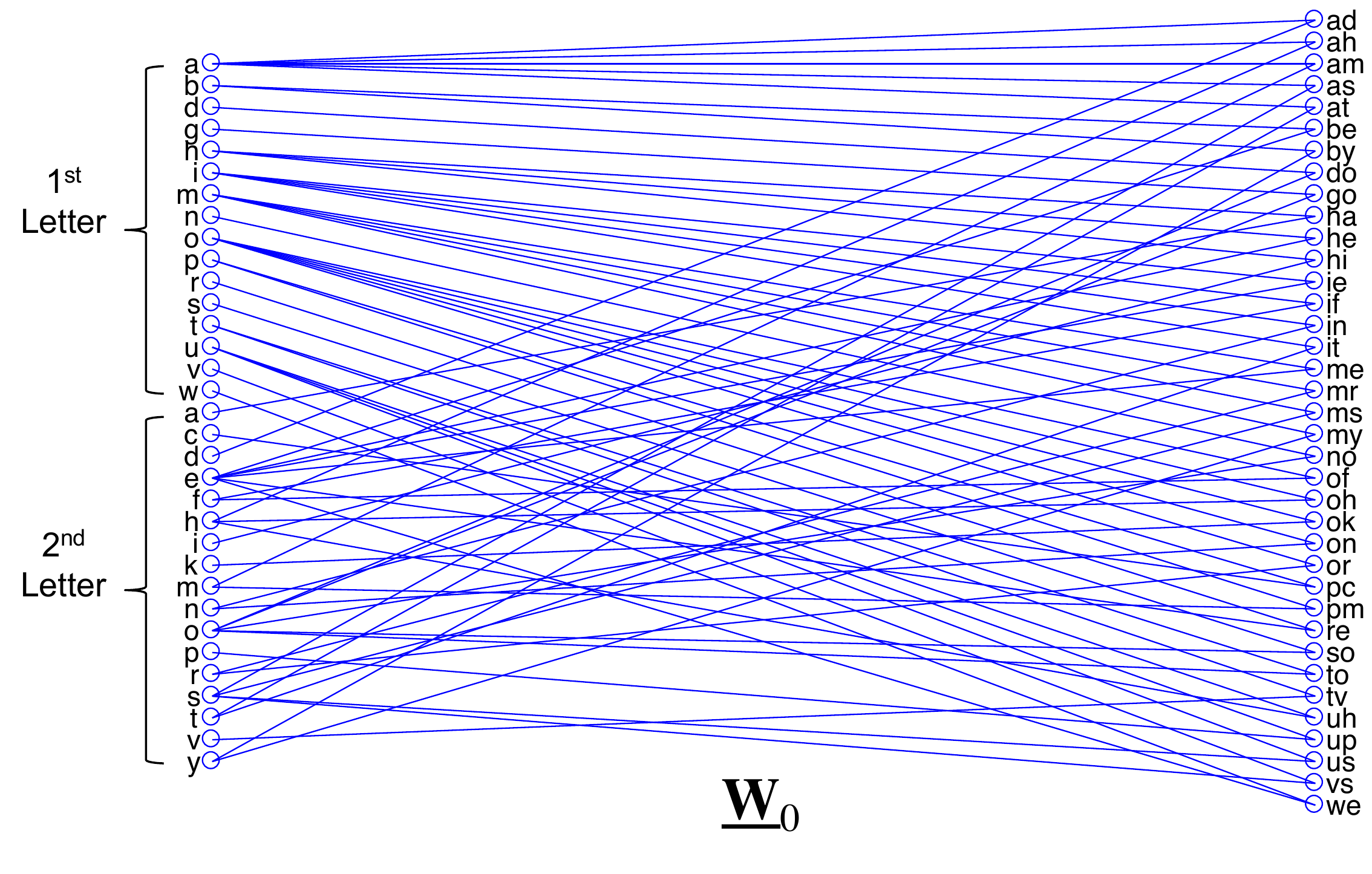}
	\caption{Single layer exact DNN solution over categories of popular two letter words.}
      	\label{fig:SelectiveSingleLayer}
\end{figure}

The single layer exact solution (see Figure~\ref{fig:SelectiveSingleLayer}) can be computed via 
$$
  \underline{\mathbf{W}}_0 = \mathbb{I}({\rm f}_{12},{\rm f}_{\underline{1}}) \oplus \mathbb{I}({\rm f}_{12},{\rm f}_{\underline{2}})
$$
where f$_{\underline{1}}$ and f$_{\underline{2}}$ are the ordered lists of the first and second letters

f$_{\underline{1}}$ = (a,a,a,a,a,b,b,d,g,h,h,h,i,i,i,i,m,m,m,m,n,

~~~~~~ o,o,o,o,o,p,p,r,s,t,t,u,u,u,v,w)

f$_{\underline{2}}$ = (d,h,m,s,t,e,y,o,o,a,e,i,e,f,n,t,e,r,s,y,o,f,h,

~~~~~~ k,n,r,c,m,e,o,o,v,h,p,s,s,e)

\begin{figure}[htb]
  	\centering
	    \includegraphics[width=\columnwidth]{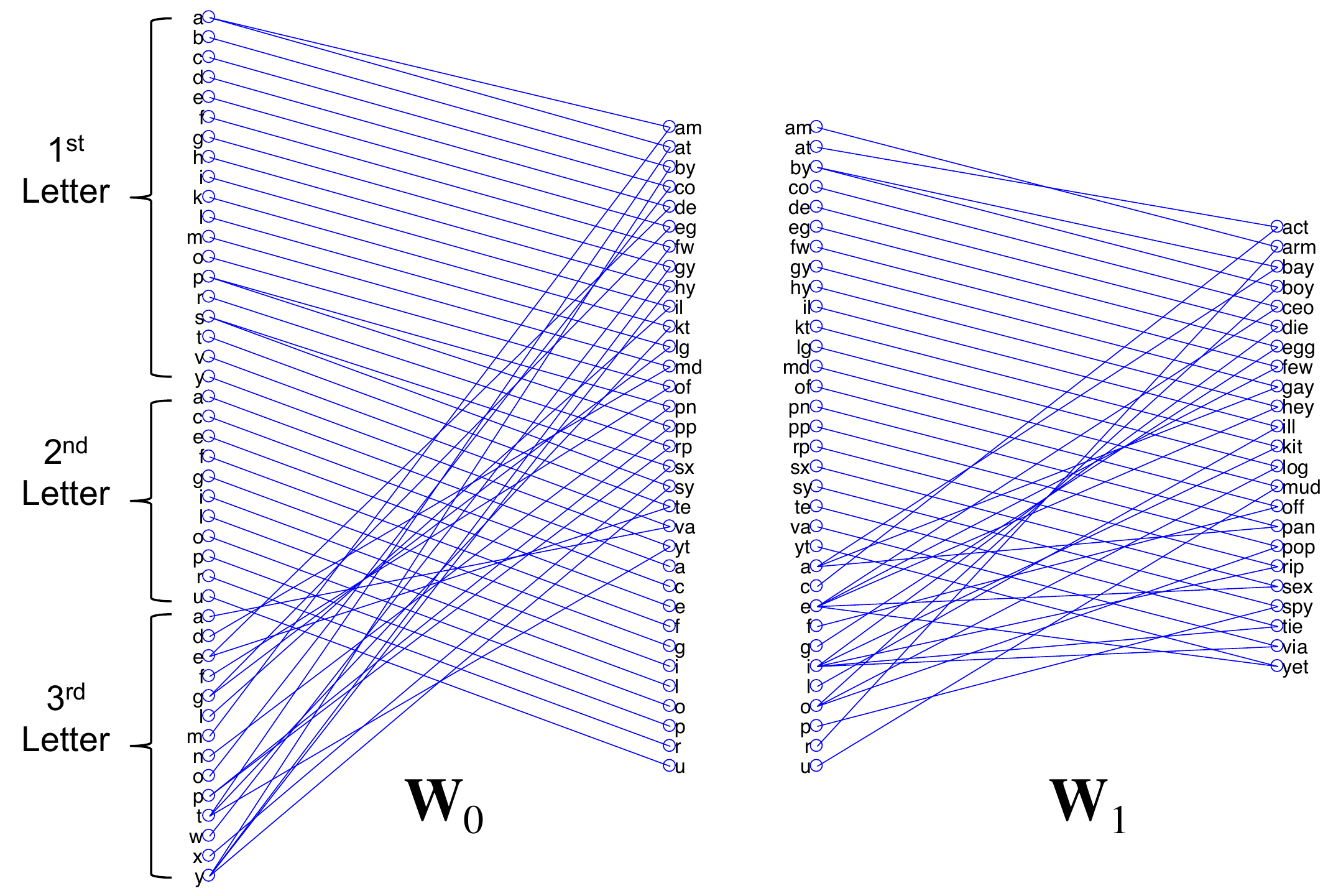}
	\caption{One of many possible two layer exact DNN solutions over categories of popular three letter words.}
      	\label{fig:SelectiveDualLayer}
\end{figure}

\begin{figure}[htb]
  	\centering
	    \includegraphics[width=\columnwidth]{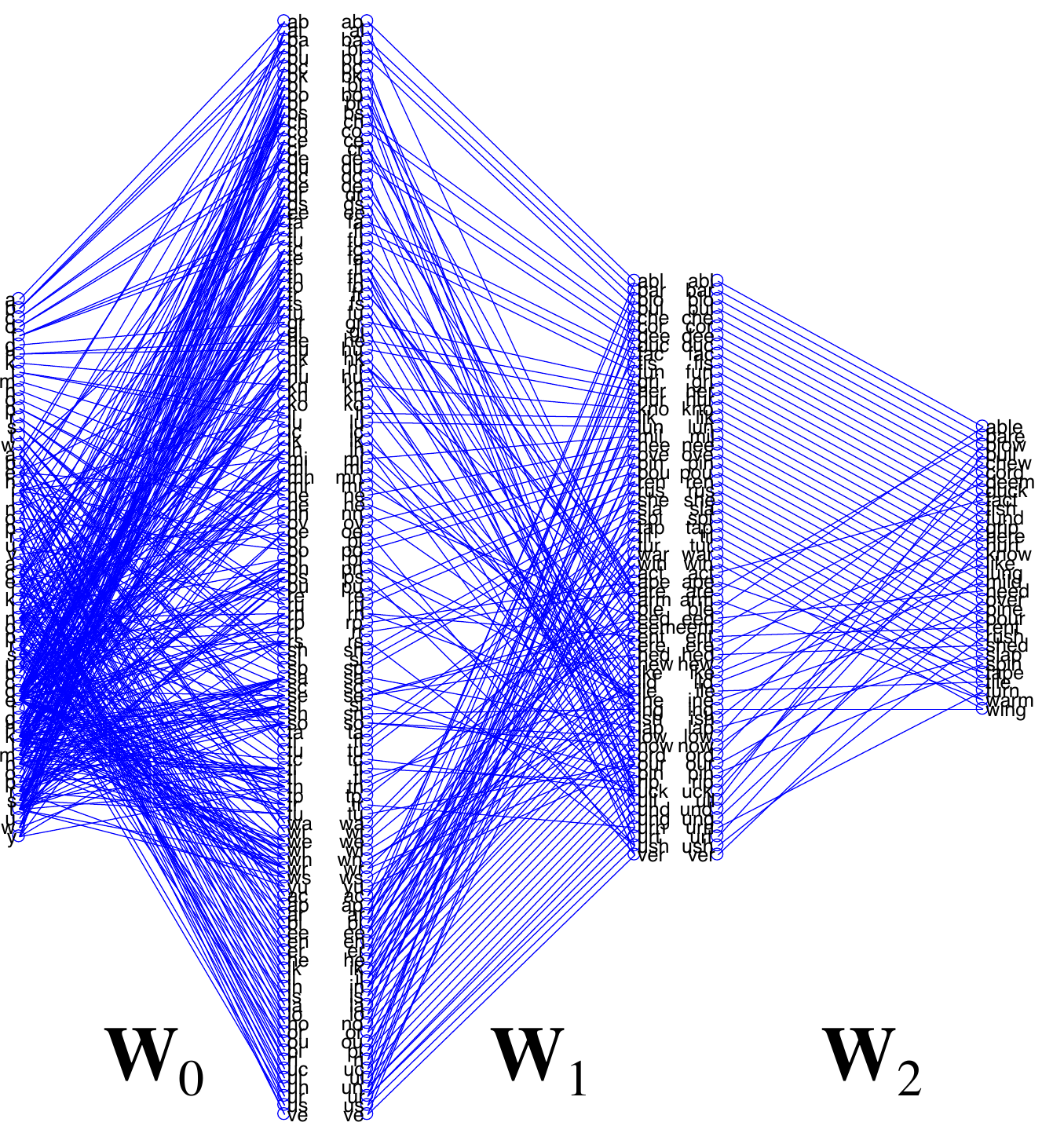}
	\caption{One of many possible three layer exact DNN solutions over categories of popular four letter words.}
      	\label{fig:SelectiveTriLayer}
\end{figure}

\noindent Similar DNN exact solutions can be selectively constructed from arbitrary sets of target categories.  Figure~\ref{fig:SelectiveDualLayer} shows one of many possible solutions for popular three letter words.
Figure~\ref{fig:SelectiveTriLayer} shows one of many possible solutions for popular four letter words.

More generally, given a set of known features f$_1$, f$_2$, f$_3$, f$_4$, and known categories f$_{1234}$, the unique single layer solution can be constructed via
$$
  \underline{\mathbf{W}}_0 = \mathbb{I}({\rm f}_{1234},{\rm f}_{\underline{1}}) \oplus \mathbb{I}({\rm f}_{1234},{\rm f}_{\underline{2}}) \oplus \mathbb{I}({\rm f}_{1234},{\rm f}_{\underline{3}}) \oplus \mathbb{I}({\rm f}_{1234},{\rm f}_{\underline{4}})
$$
One of many two layer solutions can be constructed via
\begin{eqnarray*}
\mathbf{W}_0 &=& \mathbb{I}({\rm f}_{\underline{12}},{\rm f}_{\underline{1}}) \oplus \mathbb{I}({\rm f}_{\underline{12}},{\rm f}_{\underline{2}}) \oplus \mathbb{I}({\rm f}_{\underline{34}},{\rm f}_{\underline{3}}) \oplus \mathbb{I}({\rm f}_{\underline{34}},{\rm f}_{\underline{4}}) \\
\mathbf{W}_1 &=& \mathbb{I}({\rm f}_{1234},{\rm f}_{\underline{12}}) \oplus \mathbb{I}({\rm f}_{1234},{\rm f}_{\underline{34}})
\end{eqnarray*}
Likewise, one of many three layer solutions can be constructed via
\begin{eqnarray*}
\mathbf{W}_0 &=& \mathbb{I}({\rm f}_{\underline{12}},{\rm f}_{\underline{1}}) \oplus \mathbb{I}({\rm f}_{\underline{23}},{\rm f}_{\underline{2}}) \oplus \mathbb{I}({\rm f}_{\underline{23}},{\rm f}_{\underline{3}}) \oplus \mathbb{I}({\rm f}_{\underline{34}},{\rm f}_{\underline{4}}) \\
\mathbf{W}_1 &=& \mathbb{I}({\rm f}_{\underline{123}},{\rm f}_{\underline{12}}) \oplus \mathbb{I}({\rm f}_{\underline{123}},{\rm f}_{\underline{23}}) \oplus \mathbb{I}({\rm f}_{\underline{234}},{\rm f}_{\underline{23}}) \oplus \mathbb{I}({\rm f}_{\underline{234}},{\rm f}_{\underline{34}}) \\
\mathbf{W}_2 &=& \mathbb{I}({\rm f}_{1234},{\rm f}_{\underline{123}}) \oplus \mathbb{I}({\rm f}_{1234},{\rm f}_{\underline{234}})
\end{eqnarray*}
Note: values greater than 1 are set 1 using the zero norm $| ~ |_0$ or by performing the above array addition using $\oplus = {\max}$.

As the number of layers increases, the number of possible exact DNN solutions increases combinatorially.  Most combinations of features and layers produce valid exact solutions.  To the extent that it is possible to extrapolate from these DNNs to real-world DNNs, this exponential growth might provide some insight as to why larger DNNs work.  Specifically, the computational costs of training a DNN grows approximately linearly with the number of layers.  If the potential number of equivalently good solutions grows combinatorially with the number of layers, then the probability of finding one of these solutions increases much more rapidly than the computation time.

\section{Perturbation Theory}

An important benefit of exact solutions is the ability to analyze small perturbations to gain further insight into the underlying system.  A simple perturbation is to change one of the input values in $\mathbf{y}_0$ from 1 to $r$ where $0 \leq r \leq 2$.  Varying $r$ allows a wide range of perturbations to be explored

Deletion: a correct feature is dropped ($r = 0$).

Degradation: a correct feature is reduced ($0 < r < 1$).

Normal: a correct feature is unchanged ($r = 1$).

Enhancement: a correct feature is increased ($1 < r < 2$).

Excitement:  1 is added to a correct feature ($r = 2$).

\noindent Another perturbation that can be explored is the $\beta_\ell$ that sets the size of the gap in the biases $\mathbf{b}_\ell$.  A large gap ($\beta_\ell = 1$) will reduce the triggering of neurons when incorrect features are erroneously present.  However, a large gap will also reduce the triggering of neurons if correct features are slightly degraded.  Varying $\beta_\ell$ over the range $0 < \beta_\ell \leq 1$ can be used to further explore the impact of these perturbations.

\begin{figure}[htb]
  	\centering
	    \includegraphics[width=0.8\columnwidth]{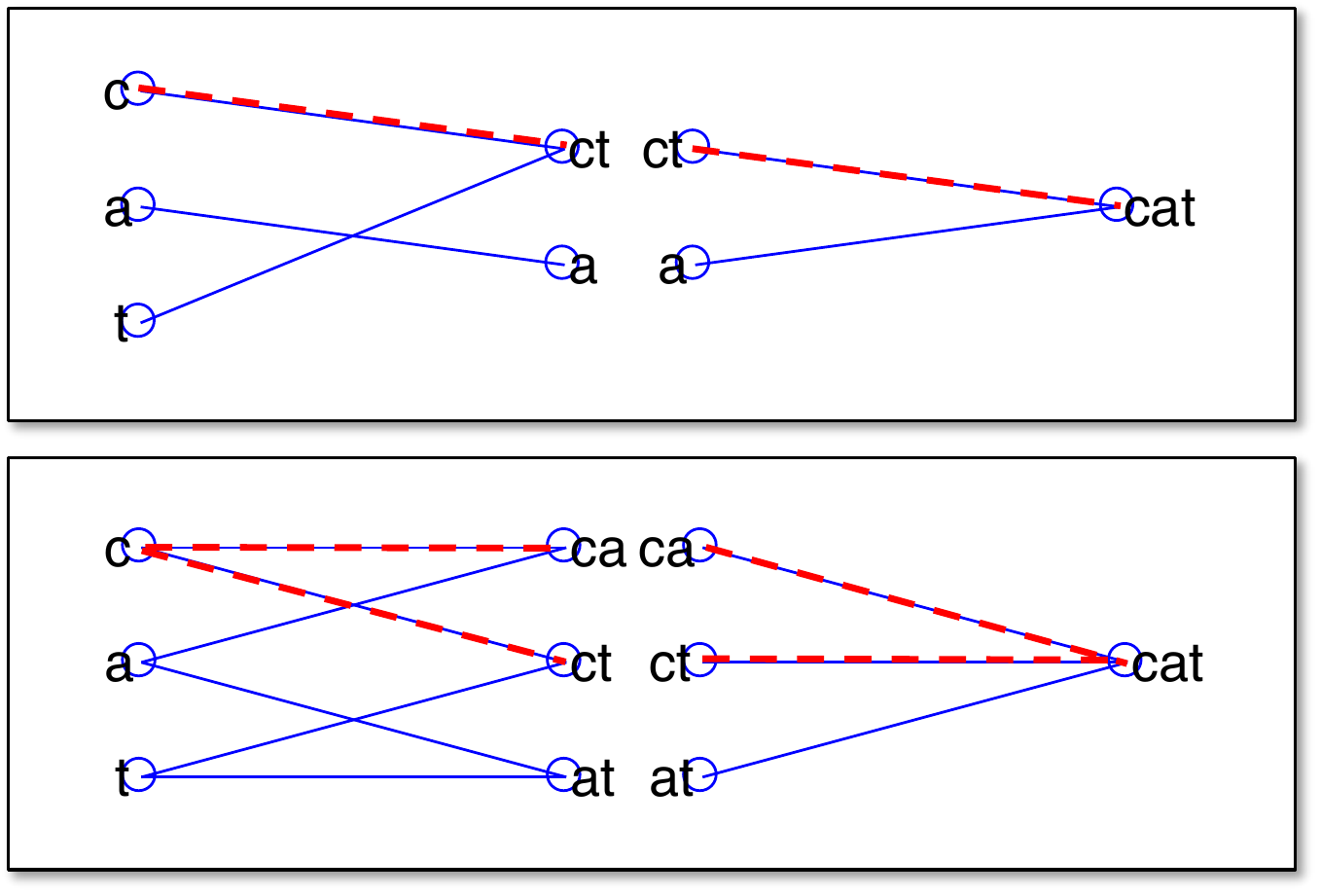}
	\caption{Supporting sub-DNNs for category $c = {\rm cat}$ are shown in black for various DNNs.  The sub-DNN depending on the feature ${\rm f} = c$ is shown in dashed red.}
      	\label{fig:PerturbationPath}
\end{figure}

The impact of perturbations to $\mathbf{y}_0$ and $\mathbf{b}_\ell$ can be measured in terms of the probability of the correct category having the maximum value in the DNN output $\mathbf{y}_L$.  In other words
$$
  {\rm arg}\max_c ~ \mathbf{y}_L(c)
$$
Given a category $c$, the sub-DNN supporting $c$ is given by
$$
  ^c\mathbf{W}_\ell = \mathbf{W}_\ell({\rm col}(^c\mathbf{W}_{\ell+1}),:) ~~~~~~~~~~ ^c\mathbf{W}_L = \mathbf{W}_L(c,:)
$$
where col() are column keys of the nonzero rows of an associative array.   Given a perturbed feature f, the sub-DNN supporting $c$ impacted by f is
$$
  ^f\mathbf{W}_{\ell+1} = ^c\mathbf{W}_{\ell+1}(:, {\rm row}(^f\mathbf{W}_\ell)) ~~~~~~~~~ ^f\mathbf{W}_0 = ^c\mathbf{W}_L(:,f)
$$
where row() are row keys of the nonzero rows of an associative array.  Examples of these supporting sub-DNNs are shown in Figure~\ref{fig:PerturbationPath}.

\begin{figure}[htb]
  	\centering
	    \includegraphics[width=\columnwidth]{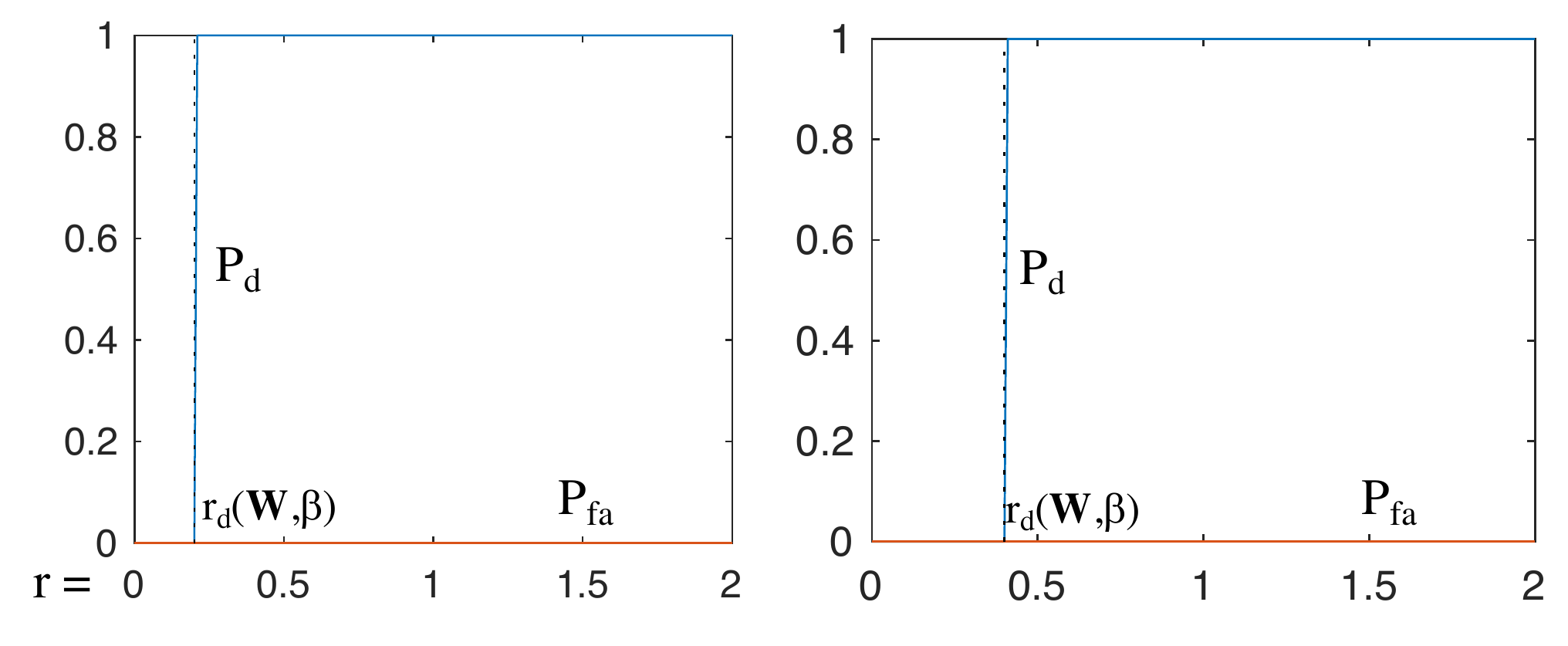}
	\caption{Plots of $P_{\rm d}(r)$ in blue and $P_{\rm fa}(r)$ in red for two DNNs. (left) Single layer ($L = 1$) DNN with $\beta = 0$. (right) Dual layer ($L = 2$) DNN with $\beta_0 = 0.8$ and $\beta_1 = 1$.}
      	\label{fig:PdPfa}
\end{figure}

The probability of correct detection $P_{\rm d}$ for these perturbations is
$$
   P_{\rm d}(r) = \left\{
     \begin{array}{lcr}
       1 & : & r > r_d(\mathbf{W},\beta)\\
       0 & : & r \le r_d(\mathbf{W},\beta)
     \end{array}
   \right.\\
 $$
The formula $r_d(\mathbf{W},\beta)$ is derived in Appendix B.  Impressively, the probability of false alarm $P_{\rm fa}(r)$ is 0 for all values of $r$, which may provide some insight into the robustness of DNNs in real-world applications.  Plots of $P_{\rm d}(r)$ and $P_{\rm fa}(r)$  for two DNNs are shown in Figure~\ref{fig:PdPfa}.  This perturbation analysis provides a convenient mechanism for exactly checking real-word DNN systems.

\section{Conclusions}

Deep neural networks (DNNs) have emerged as a key enabler of machine learning. Applying larger DNNs to more diverse applications is an important challenge. The computations performed during DNN training and inference are dominated by operations on the weight matrices describing the DNN.  As DNNs incorporate more layers and more
neurons per layers, these weight matrices may be required to be sparse because of memory limitations.  Sparse DNNs are one possible approach, but the underlying theory is in the early stages of development and presents a number of challenges, including determining the accuracy inference and selecting nonzero weights for training.   Associative array algebra has been developed by the big data community to combine and extend database, matrix, and graph/network concepts for use in large, sparse data problems. Applying this mathematics to DNNs simplifies the formulation of DNN mathematics and reveals that DNNs are linear over oscillating semirings.  This work employs associative array DNNs to construct exact solutions, and corresponding perturbation models to the rectified linear unit (ReLU) DNN equations can be used to construct test vectors for sparse DNN implementations over various precisions.  These solutions can be used for DNN verification, theoretical explorations of DNN properties, and a starting point for the  challenge of sparse training.  Interestingly, the number of exact solutions for a given DNN problem grows combinatorially with the number of DNN layers, while the computation time grows linearly.  This observation suggests that larger DNNs may perform better because the probability of finding a reasonable solution rapidly increases as the number of layers increases.

\section*{Acknowledgments}

The authors wish to acknowledge the following individuals for their contributions and support: Alan Edelman, Charles Leiserson, Doug Reynolds, Steve Pritchard, Michael Wright, Bob Bond, Dave Martinez, Sterling Foster, Paul Burkhardt, and Victor Roytburd.




\bibliographystyle{ieeetr}

\bibliography{References}

%
%
%

\section*{Appendix A: Associative Array Algebra}

The essence of associative array algebra is three operations: element-wise addition (neural network combine), element-wise multiplication (neural network intersection), and array multiplication (neural network propagation).  In brief, an associative array $\mathbf{A}$ is defined as a mapping from sets of keys to values
$$
  \mathbf{A}: K_1 \times K_2 \to \mathbb{V}
$$
where $K_1$ are the row keys and $K_2$ are the column keys and can be any sortable set, such as integers, real numbers, and strings.   In a DNN, the row keys and column can be used to clearly label features, neurons, batches, and categories.  $\mathbb{V}$ is the set of values that form a semiring $(\mathbb{V},\oplus,\otimes,0,1)$ with addition operation $\oplus$, multiplication operation $\otimes$, additive identity/multiplicative annihilator 0, and multiplicative identity 1.  Associative arrays allow arbitrary key labels for neurons which greatly simplifies the book keeping of DNN solutions, and so the number of neurons in a layer and even the number of layers need not be specified.  $\mathbb{V}$ is the set of values that form a semiring $(\mathbb{V},\oplus,\otimes,0,1)$ with addition operation $\oplus$, multiplication operation $\otimes$, additive identity/multiplicative annihilator 0, and multiplicative identity 1. The values can take on many forms, such as numbers, strings, and sets. One of the most powerful features of associative arrays is that addition and multiplication can be a wide variety of operations.  Some of the common combinations of addition and multiplication operations that have proven valuable are standard arithmetic addition and multiplication ${+}.{\times}$, union and intersection ${\cup}.{\cap}$ that are the basis of database relational algebra \cite{codd1970relational,maier1983theory,jananthan2017polystore}, and various tropical algebras that are important in finance \cite{klemperer2010product,baldwin2016understanding,masontropical} and neural networks \cite{kepner2017enabling}: ${\max}.{+}$, ${\min}.{+}$, ${\max}.{\times}$, ${\min}.{\times}$, ${\max}.{\min}$, and ${\min}.{\max}$.

The construction of an associative array is denoted
$$
  \mathbf{A} = \mathbb{A}(\mathbf{k}_1,\mathbf{k}_2,\mathbf{v})
$$
where $\mathbf{k}_1$, $\mathbf{k}_1$, $\mathbf{v}$ are vectors of the row keys, column keys, and values of the nonzero elements of $\mathbf{A}$.  In the case when the values are 1 and there is only one nonzero entry per row or column, this associative array is denoted
$$
  \mathbb{I}(\mathbf{k}_1,\mathbf{k}_2) = \mathbb{A}(\mathbf{k}_1,\mathbf{k}_2,1)
$$
and when $\mathbb{I}(\mathbf{k}) = \mathbb{I}(\mathbf{k},\mathbf{k})$, this array is referred to as the identity.

Given associative arrays $\mathbf{A}$, $\mathbf{B}$, and $\mathbf{C}$, element-wise addition is denoted
$$
   \mathbf{C} = \mathbf{A} \oplus \mathbf{B}
$$
or more specifically
$$
   \mathbf{C}(k_1,k_2) = \mathbf{A}(k_1,k_2) \oplus \mathbf{B}(k_1,k_2)
$$
where $k_1 \in K_1$ and $k_2 \in K_2$. Similarly, element-wise multiplication is denoted
$$
   \mathbf{C} = \mathbf{A} \otimes \mathbf{B}
$$
or more specifically
$$
   \mathbf{C}(k_1,k_2) = \mathbf{A}(k_1,k_2) \otimes \mathbf{B}(k_1,k_2)
$$
Array multiplication combines addition and multiplication and is written
$$
   \mathbf{C} = \mathbf{A} \mathbf{B} = \mathbf{A} {\oplus}.{\otimes} \mathbf{B}
$$
or more specifically
$$
   \mathbf{C}(k_1,k_2) = \bigoplus_\ell \mathbf{A}(k_1,k) \otimes \mathbf{B}(k,k_2)
$$
where $k$ corresponds to the column key of $\mathbf{A}$ and the row key of $\mathbf{B}$. Finally, the array transpose is denoted
$$
   \mathbf{A}(k_2,k_1) = \mathbf{A}^{\sf T}(k_1,k_2)
$$
The Kronecker product of two arrays
$$
{\bf C} = {\bf A} \mathbin{\text{\textcircled{$\otimes$}}} {\bf B}
$$
is defined as follows
  \[
    {\bf C}\bigl(k_1k_3,k_2k_4) = {\bf A}(k_1,k_2) \otimes {\bf B}(k_3,k_4)
  \]
where $k_1k_2$ and $k_3k_4$ are order preserving combinations of $k_1$ and $k_2$ and $k_3$ and $k_4$, respectively.

The above operations have been found to enable a wide range of algorithms and matrix mathematics while also preserving several valuable mathematical properties that ensure the correctness of parallel execution.  These properties include commutativity
\begin{eqnarray*}
   \mathbf{A} \oplus \mathbf{B} &=& \mathbf{B} \oplus \mathbf{A} \\
   \mathbf{A} \otimes \mathbf{B} &=& \mathbf{B} \otimes \mathbf{A} \\
   (\mathbf{A} \mathbf{B})^{\sf T} &=& \mathbf{B}^{\sf T} \mathbf{A}^{\sf T}
\end{eqnarray*}
associativity
\begin{eqnarray*}
   (\mathbf{A} \oplus \mathbf{B}) \oplus \mathbf{C} &=& \mathbf{A} \oplus (\mathbf{B} \oplus \mathbf{C}) \\
   (\mathbf{A} \otimes \mathbf{B}) \otimes \mathbf{C} &=& \mathbf{A} \otimes (\mathbf{B} \otimes \mathbf{C}) \\
   (\mathbf{A}  \mathbf{B})  \mathbf{C} &=& \mathbf{A}  (\mathbf{B}  \mathbf{C}) \\
   (\mathbf{A} \mathbin{\text{\textcircled{$\otimes$}}} \mathbf{B}) \mathbin{\text{\textcircled{$\otimes$}}} \mathbf{C} &=& \mathbf{A} \mathbin{\text{\textcircled{$\otimes$}}} (\mathbf{B} \mathbin{\text{\textcircled{$\otimes$}}} \mathbf{C})
\end{eqnarray*}
distributivity
\begin{eqnarray*}
   \mathbf{A} \otimes (\mathbf{B} \oplus \mathbf{C}) &=& (\mathbf{A} \otimes \mathbf{B}) \oplus (\mathbf{A} \otimes \mathbf{C}) \\
   \mathbf{A} (\mathbf{B} \oplus \mathbf{C}) &=& (\mathbf{A}  \mathbf{B}) \oplus (\mathbf{A}  \mathbf{C})  \\
   \mathbf{A} \mathbin{\text{\textcircled{$\otimes$}}} (\mathbf{B} \oplus \mathbf{C}) &=& (\mathbf{A} \mathbin{\text{\textcircled{$\otimes$}}} \mathbf{B}) \oplus (\mathbf{A} \mathbin{\text{\textcircled{$\otimes$}}} \mathbf{C})
\end{eqnarray*}
and the additive and multiplicative identities
$$
   \mathbf{A} \oplus \mathbs{0} = \mathbf{A} ~~~~~~~~~~~~ \mathbf{A} \otimes \mathbs{1} = \mathbf{A} ~~~~~~~~~~~~ \mathbf{A} \mathbb{I} = \mathbf{A}
$$
where $\mathbs{0}$ is an array of all 0, $\mathbs{1}$ is an array of all 1, and $\mathbb{I}$ is an array with 1 along its diagonal.  Furthermore, these arrays possess a multiplicative annihilator
$$
   \mathbf{A} \otimes \mathbs{0} = \mathbs{0} ~~~~~~~~~~~~ \mathbf{A} \mathbs{0} = \mathbs{0}
$$
~~~~ 

\section*{Appendix B: Perturbation Propagation}

Correct category is the top category and passes through sub-DNN unaffected by $h()$, making the sub-DNN for this category a linear equation
$$
  \mathbf{y}_{\ell+1} = ~ ^c\mathbf{W}_\ell \mathbf{y}_\ell + \mathbf{b}_\ell
$$
where
$$
  \mathbf{b}_{\ell} = \beta_\ell \mathbf{1} - ^c\mathbf{W}_\ell \mathbf{1}
$$
Inserting the above expression into the DNN equations yields
\begin{eqnarray*}
\mathbf{y}_L &=& \left( \prod_{\ell=L-1}^{0} ~ ^c\mathbf{W}_\ell \right) \mathbf{y}_0 + \left( \prod_{\ell=L-1}^{0} ~ ^c\mathbf{W}_\ell \right) \mathbf{b}_0 + \\
             && \sum_{j=2}^{L-1} \left( \prod_{\ell=L-1}^{j} ~ ^c\mathbf{W}_\ell \right) \mathbf{b}_{j-1}
\end{eqnarray*}
Substituting
$$
  \mathbf{b}_0 = \beta_0 \mathbf{1} - ^c\mathbf{W}_0 \mathbf{1}
$$
into the prior expression gives
\begin{eqnarray*}
\mathbf{y}_L &=& \left( \prod_{\ell=L-1}^{0} ~ ^c\mathbf{W}_\ell \right) (\mathbf{y}_0 - \mathbf{1})  + \beta_0 \left( \prod_{\ell=L-1}^{0} ~ ^c\mathbf{W}_\ell \right) \mathbf{1} + \\
             && \sum_{j=2}^{L-1} \left( \prod_{\ell=L-1}^{j} ~ ^c\mathbf{W}_\ell \right) \mathbf{b}_{j-1}
\end{eqnarray*}
The input for feature f is
$$
  \mathbf{y}_0 = \mathbf{1} - (r-1) \mathbb{I}({\rm f})
$$
resulting in
\begin{eqnarray*}
\mathbf{y}_L &=& (r-1) \left( \prod_{\ell=L-1}^{0} ~ ^c\mathbf{W}_\ell \right) \mathbb{I}({\rm f})  + \beta_0 \left( \prod_{\ell=L-1}^{0} ~ ^c\mathbf{W}_\ell \right) \mathbf{1} + \\
             && \sum_{j=2}^{L-1} \left( \prod_{\ell=L-1}^{j} ~ ^c\mathbf{W}_\ell \right) \mathbf{b}_{j-1}
\end{eqnarray*}
The correct top category will be selected when $\mathbf{y}_L > 0$ or $r > r_d(\mathbf{W},\beta)$, resulting in 
$$
r_d(\mathbf{W},\beta) = 1 - \frac{\beta_0 \left( \prod_{\ell=L-1}^{0} ~ ^c\mathbf{W}_\ell \right) \mathbf{1} + 
             \sum_{j=2}^{L-1} \left( \prod_{\ell=L-1}^{j} ~ ^c\mathbf{W}_\ell \right) \mathbf{b}_{j-1}}{\left( \prod_{\ell=L-1}^{0} ~ ^c\mathbf{W}_\ell \right) \mathbb{I}({\rm f})}
$$
\end{document}